\pdfoutput=1
\PassOptionsToPackage{sort}{natbib}  
\documentclass{article}

\usepackage[preprint,nonatbib]{neurips_2021}

\usepackage[utf8]{inputenc} 
\usepackage[T1]{fontenc}    
\usepackage{url}            
\usepackage{booktabs}       
\usepackage{amsfonts}       
\usepackage{nicefrac}       
\usepackage{microtype}      
\usepackage{xcolor}         
\usepackage{amsmath}

\usepackage{mathtools}

\usepackage{subcaption}
\usepackage{booktabs}
\usepackage{graphicx}
\usepackage{wrapfig}
\usepackage[pagebackref=true,breaklinks=true,colorlinks,citecolor=blue,linkcolor=blue,bookmarks=false]{hyperref}
\usepackage{paralist}
\usepackage{algorithm}
\usepackage{algorithmic}

\title{Learning 3D Dense Correspondence via\\ Canonical Point Autoencoder}

\author{
An-Chieh Cheng$^{1}$, 
 \textbf{Xueting Li$^{2}$}, \
 \textbf{Min Sun$^{1}$}, \
 \textbf{Ming-Hsuan Yang$^{2}$}, 
 \textbf{Sifei Liu$^{3}$}, \
\\ $^{1}$National Tsing-Hua University \
 $^{2}$University of California, Merced, \
 $^{3}$NVIDIA}

\begin{document}

\maketitle

\pdfoutput=1
\begin{abstract}
We propose a canonical point autoencoder (CPAE) that predicts dense correspondences between 3D shapes of the same category.
The autoencoder performs two key functions: (a) encoding an arbitrarily ordered point cloud to a canonical primitive, e.g., a sphere, and (b) decoding the primitive back to the original input instance shape.
As being placed in the bottleneck, this primitive plays a key role to map all the unordered point clouds on the canonical surface and to be reconstructed in an ordered fashion.
Once trained, points from different shape instances that are mapped to the same locations on the primitive surface are determined to be a pair of correspondence.
%
Our method does not require any form of annotation or self-supervised part segmentation network and can handle unaligned input point clouds.
Experimental results on 3D semantic keypoint transfer and part segmentation transfer show that our model performs favorably against state-of-the-art correspondence learning methods. 
\end{abstract}
\pdfoutput=1
\vspace{-3mm}
\section{Introduction}
\vspace{-1mm}
\label{section:introduction}

With prior knowledge and experience, humans can easily perceive corresponding object parts (e.g., the wings from two different airplanes), understand their shape and appearance variance, in order to distinguish different objects coming from the same category.
In computer vision, modeling dense correspondence between 3D shapes in one category is fundamental for numerous applications, such as robot grasping~\cite{saxena2007robotic,miller2003automatic}, object manipulation~\cite{hao2013efficient} and texture mapping~\cite{zeng20203d,miller2003automatic}.
However, existing 3D cameras typically capture raw point clouds of shape surfaces that are arbitrarily-ordered and unstructured, in which correspondences are not established. 
3D mesh representation, although is usually parameterized with UV maps that can indicate correspondences, cannot be directly obtained from sensors and needs to be reconstructed from other types of representations, e.g., 2D images \cite{umr2020,cmrKanazawa18} or 3D point clouds \cite{badki2020meshlet}. 
In this work, we focus on learning point cloud correspondences, which remains an open challenge since it is infeasible to label ground truth correspondence annotations.


Without ground truth annotations, existing methods mainly discover shape correspondences via seeking a form of canonical space that can associate various instance shapes.
For example, in particular shape domains such as human bodies~\cite{groueix20183d} and human faces~\cite{gilani2017dense}, parameterized shape primitives have been designed to fit the observed raw data and to obtain the correspondences. Such designs, however, cannot be generalized to other categories, e.g., man-made objects~\cite{chang2015shapenet}.
%
Recently, several part co-segmentation networks relax the requirements of specific parameterized primitives, but instead decompose input shapes into an ordered group of simplest part constitutions~\cite{chen2019bae,liu2020learning}, in a self-supervised manner.
These methods, however, require careful selection of the autoencoder architectures (i.e., they need to be considerably shallow to let the branches only able to represent simple shapes), and the number of part bases.  
%
Moreover, such part-based representation does not explicitly provide fine-grained (e.g., point-level) correspondences.


In this work, we introduce a novel canonical space where dense (i.e., point-level) correspondences for all the shapes of a category can be explicitly obtained from.
Inspired by 3D mesh representation~\cite{kanazawa2018learning,ucmrGoel20,tulsiani2020implicit,umr2020} where shapes from one category are represented as deformations on top of a shape primitive, in our work, 
we set the canonical space as a 3D UV sphere.
Our goal is to learn a ``point cloud-to-sphere mapping'' such that corresponding parts from different instances overlap when mapped onto the canonical sphere.
In other words, similar to the mesh representation, a unique UV coordinate can represent the same semantic point/local region of shapes (e.g., the tip of an aeroplane's wing), regardless of shape variations.
Towards this goal, we introduce the canonical point autoencoder (CPAE):
we place the sphere primitive at the bottleneck; the encoder non-linearly maps each individual input shape to the sphere primitive, where the decoder deforms the primitive back to match the original shape.
We show that with several self-supervised objectives, this autoencoder architecture effectively (1) enforces the input points warped to the surface of the sphere primitives, and (2) encourages those corresponding points from different instances mapped to the same location on the sphere -- both guarantee that the network learns correct dense correspondences.
Essentially, we \textit{do not} assume all object shapes in one category having the same topology, e.g., an armchair does not have correspondence on its armrests, with another instance without an armrest. To introduce such uncertainty for correspondence matching, we propose an adaptive Chamfer loss on the bottleneck to allow customized primitive for each instance. 
As such, we are able to determine if a point on one instance has a correspondence in another point cloud. 

One advantage of the proposed method compared to the recent work~\cite{liu2020learning} is that we can learn correspondences even when instances in the training dataset are not aligned, i.e., our model is rotation-invariant within a certain rotation range and does not need to predict an additional rotation matrix as used in~\cite{liu2020learning}. 
The main contributions of this work are:
\begin{compactitem}
    \item We introduce a novel canonical space -- a UV sphere, that explicitly represents dense correspondences of shapes from one category.
    \item With the canonical space on the bottleneck, we design an autoencoder that learns such a ``point cloud-to-sphere mapping'' via a group of self-supervised objectives.
    \item We apply the proposed method on various categories and quantitatively evaluate on the task of 3D semantic keypoint transfer and part segmentation label  transfer, achieving comparable if not better performance than state-of-the-art methods.
\end{compactitem}

\pdfoutput=1
\vspace{-2mm}
\section{Related Work}
\label{section:relatedwork}

\vspace{-2mm}
\paragraph{Deep Learning on Point Clouds.} 
As a flexible and memory efficient representation of 3D shapes, point cloud has been widely studied and combined with deep neural networks.
The PointNet~\cite{qi2017pointnet} solves point cloud classification and segmentation by using MLPs and a max pooling layer to aggregate 3D shape information. One crucial property of the PointNet is that it is able to handle unordered input and is thus invariant to point permutations. In our work, we utilize the first few layers of the PointNet as our shape encoder.
Another line of works~\cite{yang2018foldingnet,deprelle2019learning,groueix2018papier,yang2019} focus on reconstructing or generating point clouds.
The most representative and related to our work is the FoldingNet~\cite{yang2018foldingnet}, where a point cloud is first encoded by a graph-based encoder and then reconstructed by sequentially applying the ``folding operation'' (instantiated by MLPs) to a 2D UV map.
In this work, we reveal why the MLPs are able to preserve point order and utilize it as building blocks in our CPAE.

\vspace{-2mm}
\paragraph{3D Dense Correspondence.}
Given a pair of source and target instance in the same category, 3D dense correspondence learning targets at finding a corresponding point in the target instance for each point from the source instance.
Several approaches~\cite{chen2019edgenet,choy2019fully,gojcic2019perfect} resolve this task through point cloud registration, using labeled pairwise correspondence as supervision.
To relax constraints on supervision, Bhatnagar et al.~\cite{bhatnagar2020ipnet} predict part correspondences to a template via implicit functions. Unfortunately, they require part labels for training.
To unsupervisedly learn 3D dense correspondence, Chen et al.~\cite{chen2020unsupervised} propose a method to learn 3D structure points that are consistent across different instances. However, the model assumes structure similarity among different instances, which ignores intra-class variants and fails to detect non-existing correspondences between dissimilar shapes in the same category.
Most relevant to our work, Liu et al.~\cite{liu2020learning} introduce an unsupervised approach that leverages part features learned by the BAE-NET~\cite{chen2019bae} to build dense correspondences. 
Their algorithm is feasible to calculate a confidence score representing the probability of correspondence. 
However, in order to train the implicit function, they require additional knowledge of object surface to compute the occupancy. 
Moreover, the correspondence learning in~\cite{liu2020learning} heavily relies on the completeness of the part features from the BAE-NET~\cite{chen2019bae}, which can lead to incorrect correspondences for parts that BAE-NET cannot separate (e.g., flat surfaces, objects with fine-grained details). 
In contrast, our approach directly learns dense correspondences from the point cloud with self-supervision.

\vspace{-2mm}
\paragraph{3D Deformable Mesh Representations.}
Our algorithm is also related to methods~\cite{kanazawa2018learning,ucmrGoel20,tulsiani2020implicit,umr2020} that represent instance shapes as mesh deformations of a mesh primitive, i.e. a sphere template. 
Vertices on each instance surface that are mapped to the same locations on the shape primitive are discovered as correspondences.
One obvious limitation is that only genus 0 shapes, e.g., birds \cite{kanazawa2018learning,umr2020}  can be deformed from a mesh sphere. In contrast, ours does not have such restriction thanks to the proposed non-linear mappings, instead of explicit deformation.

\pdfoutput=1
\vspace{-1mm}
\section{Proposed Method}
\vspace{-1mm}
\label{section:methodology}

\begin{figure*}[t]
\begin{center}
\includegraphics[width=1.0\linewidth]{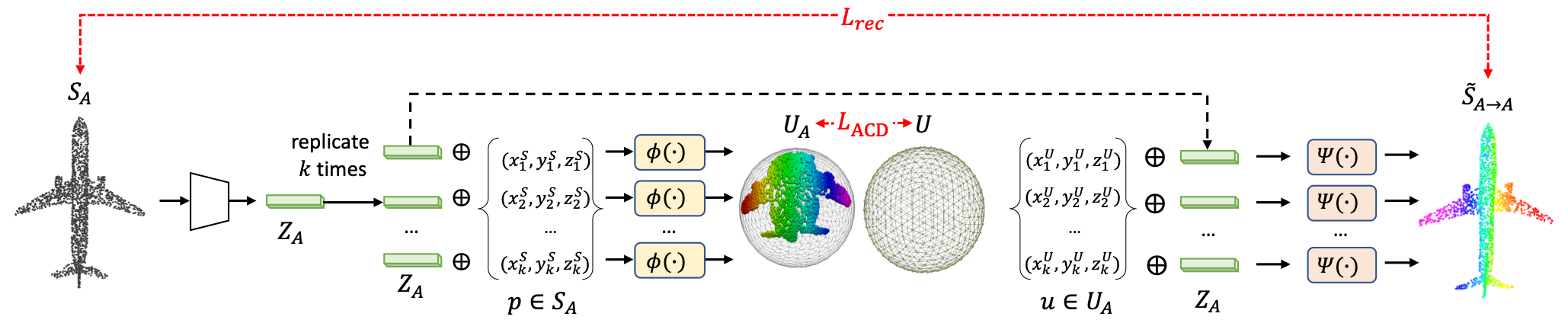}
\end{center}
\caption{Overview of the proposed method. Here, $p=(x_{i}^{S},y_{i}^{S},z_{i}^{S})$ is a point on the input point cloud $S_A$, $u=(x_{i}^{U},y_{i}^{U},z_{i}^{U})$ is a point on the primitive $U_A$. The ``$\oplus$'' sign indicates concatenation. $\Phi(\cdot)$ and $\Psi(\cdot)$ are MLPs as discussed in Section~\ref{section:Network}.}
\label{fig.overview}\vspace{-3mm}
\end{figure*}

In this section, we introduce our end-to-end CPAE for learning dense correspondences from point clouds without ground truth annotation.
Given a point cloud $S_A$ with individual point $p\in \mathbb{R}^3$ (see Figure~\ref{fig.overview}), our model: (1) predicts its canonical primitive $U_A$ (a ``deformed'' point cloud with the same number of points as $S_A$), which is supposed to be as close as possible to the canonical sphere in the bottleneck; (2) reconstructs the original input point cloud back from $U_A$.
We show that while the first ensures each input point cloud to be warped to the surface of the sphere primitive, the second indirectly encourages the corresponding points from different point cloud instances to overlap during mapping to the primitive.
In the following, we first describe our network architecture, i.e., the encoder and decoder modules.
We then introduce the adaptive Chamfer loss and the reconstruction losses that are applied to each individual module. 
Finally, we show that our decoder reconstructs ordered point clouds, i.e., different point cloud instances can fetch their correspondences directly via the point indices, which provides a more accurate inference.

\vspace{-1mm}
\subsection{Network Architecture}
\vspace{-1mm}
\label{section:Network}

Our method learns two mapping functions: one maps each individual 3D point $p\in \mathbb{R}^3$ in the world coordinates to the canonical space $\Phi(p)=u, u\in \mathbb{R}^3$, while the other conducts the inverse mapping $\Phi^{-1}(u)=p$.
Instead of using a reversible network, we instantiate $\Phi(\cdot)$ and $\Phi^{-1}(\cdot)$ with two MLPs respectively, and cascade them to construct an autoencoder (i.e., $\Phi(\cdot)$ is the encoder and $\Psi(\cdot)\approx\Phi^{-1}(\cdot)$ is the decoder).
%
However, this mapping function cannot be generalized to different point clouds if it only receives the coordinate of a single point $p$ as the input.
Therefore, we formulate them as conditional mapping functions by concatenating the input with a shape latent code $Z_A$ that represents a unique input shape $S_A$, e.g., $\Phi(p, Z_A)$, in order to generalize the mapping functions to all instances in the same category.

There are three key modules in the proposed method: (1) a PointNet encoder that produces the shape latent code; (2) an encoder MLP (denoted as \textit{canonical} mapping) that maps a point cloud to a primitive sphere, and (3) a decoder MLP (denoted as \textit{inverse} mapping) that deforms the primitive sphere back to the input point cloud. 

\vspace{-2mm}
\paragraph{PointNet Encoder.}
As shown in Figure~\ref{fig.overview}, given a point cloud $S_A\in \mathcal{R}^{k\times 3}$ with $k$ points, we encode it using a PointNet~\cite{qi2017pointnet} encoder in a way similar to ~\cite{deprelle2019learning,groueix20183d,groueix2018papier}. 
Each 3D point of the input point cloud is represented as a 512 dimensional vector using an MLP with 3 hidden layers of 64, 128, 512 neurons and ReLU activations. 
We then aggregate all point features with max-pooling and a linear layer to generate a global latent code $z_A\in \mathcal{R}^{512}$. 
%
We use PointNet because it produces a robust latent code representing the global shape, which is also invariant to input permutation. 

\vspace{-2mm}
\paragraph{Canonical Mapping $\Phi(\cdot)$.}
\label{section:methodology:unfoding}
To learn the canonical mapping function, we concatenate each point with the global latent code $[p,Z]\in \mathbb{R}^{515}$ as input to the MLP, which then outputs a 3D coordinate $u\in \mathbb{R}^{3}$ in the canonical space.
We construct a sphere point cloud by uniformly sampling a large number of points from a standard sphere mesh, and place them as the canonical primitive at the bottleneck.
A Chamfer loss is used to measure the difference between the outputted point ${u}$ and primitive to encourage the mapped point adhering to the surface  (see Eq.~\eqref{eq:unfolding}).
%

Ideally, all the mapped instances from the same category should be as close as possible to the canonical sphere primitive.  
However, an instance may include parts/regions that do not exist in other instances due to intra-class variation, e.g., not all chairs include armrests.
With a conventional Chamfer loss, the shapes for objects with rare components do not converge during training, i.e., those rare components are mapped to locations that are far away from the primitive surface. 
%
Thus, we relax the bidirectional constraint of the Chamfer loss and allow each instance to produce its own ``instance primitive'' to some extent (see $U_A$ in Figure~\ref{fig.overview} ).
This is formulated via an adaptive Chamfer loss $L_{ACD}$:
\begin{equation}
L_{ACD}(U_A, U) = \frac{1}{| U_A |}\sum_{p\in U_A}\min_{q\in U}\left\Vert p - q \right\Vert_2 + \alpha \frac{1}{| U |} \sum_{q\in U}\min_{p\in U_A}\left\Vert q - p \right\Vert_2
\label{eq:unfolding}
\end{equation}
where $\alpha \sim [0, 1]$ is an adaptive parameter indicating to what extent the predicted instance primitive should match a canonical sphere, $U_A$ and $U$ are the instance primitive and the canonical UV primitive (e.g., a 3D UV sphere), respectively.
When $\alpha=1$, $L_{ACD}$ is equivalent to calculating the conventional Chamfer distance between the unfolded primitive and the canonical UV primitive.
During training, $\alpha$ is initialized to 1 and gradually decreased to 0. 
This allows the canonical mapping to predict instance-aware primitive since the second term of the Chamfer loss no longer constraints the primitive to be consistent with a canonical sphere.
In the experiments (see Section~\ref{section:experiments:ablations}), we show that this design allows us to infer rare object components that have no correspondences in other instances since non-corresponding regions usually occupy a distinct area in the canonical space.

\vspace{-2mm}
\paragraph{Inverse Mapping $\Psi(\cdot)$.}
Similarly, we utilize another MLP in the decoder, which receives the concatenation of one point from the bottleneck and the global shape code, i.e., $[u,Z]\in \mathbb{R}^{515}$, as the input. The function learns to map it back to its original world coordinate $p$, which can be fulfilled via a point-to-point reconstruction loss. 
We leverage an MES loss $L_{MSE}$, a Chamfer distance $L_{CD}$ and an Earth Mover Distance (EMD) $L_{EMD}$ between the reconstruction $\hat{S}$ and input point cloud $P$:
\begin{equation}
L_{rec}(P,\hat{S})= \mu_{1}L_{MSE}(P, \hat{S}) + \mu_{2}L_{CD}(P, \hat{S}) + \mu_{3}L_{EMD}(P, \hat{S})
\label{eq:Rec}
\end{equation}
where $\mu_{1}, \mu_{2}, \mu_{3}$ are the weights, and empirically determined, $\mu_{1} = 1e3$, $\mu_{2} = 1e1$, and $\mu_{3} = 1$.

The inverse mapping bears some resemblance to the principle of FoldingNet~\cite{yang2018foldingnet}, which adopts an MLP architecture to map one 2D coordinate together with a global shape representation to the 3D world\footnote{Unlike our method, it utilizes and samples points from a standard 2D UV map instead of a 3D sphere.}. 
However, since FoldingNet does not have the forward mapping function as ours and the input points are untraceable, it needs to use a Chamfer loss, instead of a point-wise loss (Eq.~\eqref{eq:Rec}) as ours.

 \begin{wrapfigure}{r}{0.5\textwidth}
     \vspace{-15pt}
     \centering
     \includegraphics[width=0.48\textwidth]{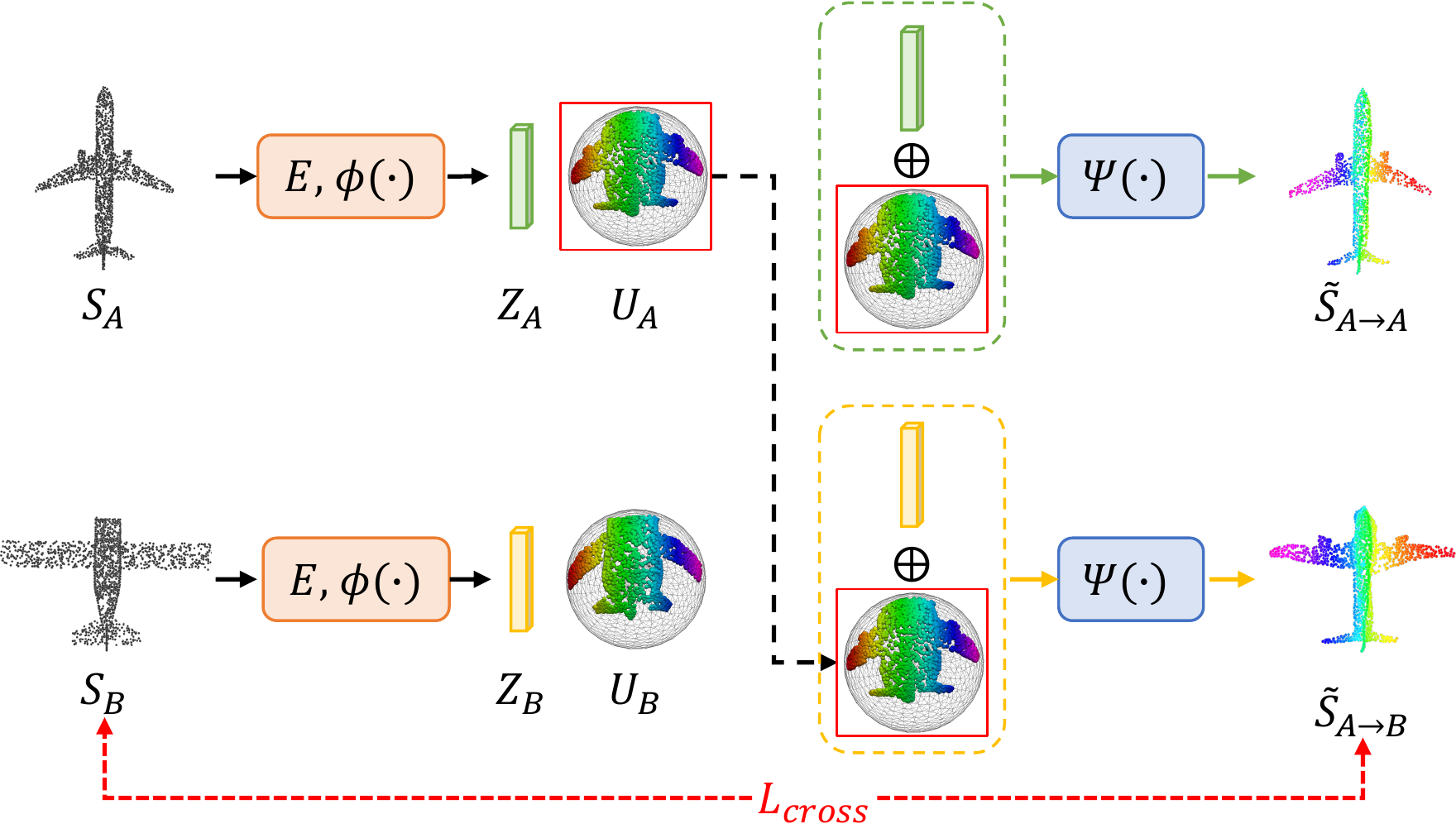}
     \caption{Cross-reconstruction. The $E, \Phi(\cdot),  \Psi(\cdot)$ are identical as in Figure~\ref{fig.overview}.}
     \label{fig.cross}
     \vspace{-15pt}
 \end{wrapfigure}

\vspace{-2mm}
\paragraph{Align Canonical Mappings via Cross-Reconstruction.}
With the encoder and objective function, the canonical mapping is able to map the input points to the surface of the canonical primitive. 
However, even with the reconstruction loss (Eq.~\eqref{eq:Rec}), there is no guarantee that the corresponding points from different shape instances can be overlapped on the sphere -- which is the key to learn dense correspondences.
For instance, in Figure~\ref{fig.overview}, while $S_A$ is mapped to the frontal half of the sphere, it is likely that another shape $S_B$ will be mapped to the rear half.

To better align the mapped points, we introduce a cross-structured decoder that leverages: (1) a self-reconstruction branch as presented in Eq.~\eqref{eq:Rec}, and (2) a cross-reconstruction branch.
As shown in Figure~\ref{fig.cross}, we feed the combination of the predicted instance primitive $U_A$ of point cloud $S_A$, with the shape latent vector $Z_{B}$ of another randomly sampled point cloud $S_B$ to the folding decoder.
We then minimize the Chamfer distance between the output shape $\hat{S}_{A \rightarrow B}$ and $S_{B}$. 
%
The cross-reconstruction loss between the point cloud $S_A$ and $S_B$ is:
\begin{equation}
L_{cross}(S_A, S_B) = L_{CD}(\hat{S}_{A \rightarrow B}, S_{B})
\label{eq:cross}
\end{equation}
%
%
The cross-reconstruction branch ensures even by swapping the predicted primitives $U_A$ and $U_B$, the inverse mapping can still reconstruct their own shapes, conditioned on their own shape latent codes.
That is, the decoder encourages $U_A$ and $U_B$ to overlap on the canonical primitive, and thus ensures the network learns correct correspondences.

\vspace{-2mm}
\paragraph{Relation to Implicit Function.}
%
We note that since both mapping functions $\Phi(\cdot)$ and $\Psi(\cdot)$ process each input point independently, they can be interpreted as conditional implicit functions, which are widely applied for 3D shape reconstruction~\cite{park2019deepsdf,mescheder2019occupancy}, view synthesis~\cite{sitzmann2019srns,mildenhall2020nerf}, and recently for self-supervised 3D correspondence \cite{liu2020learning}.
Such an interpretation reveals that the MLPs indeed learn continuous mappings that are feasible for interpolations, e.g., once the decoder MLP is learned, it is able to map any point on a continues sphere surface that is not necessarily among the mapped input ${u}$, to the world coordinate. E.g, one can sample points densely from the surface of the sphere to reconstruct a continuous surface or a 3D mesh.

\vspace{-1mm}
\subsection{Finding Correspondences from Ordered Point Clouds}
\vspace{-1mm}
\label{section:methodology:folding}

In this section, we show that the CPAE is able to generate ordered point clouds (see the output point clouds in Figure~\ref{fig.overview}). 
Compared to the methods obtaining correspondences by directly tracing overlapped points from the canonical space, our model that makes use of ordered output point cloud generates more accurate results during the inference stage. 
%


\vspace{-2mm}
\paragraph{Reconstruction of Ordered Point Clouds.}
We first show that an MLP~\cite{yang2018foldingnet} decoder preserves such correspondence, i.e. corresponding points outputted by the decoder are mapped from the same points on the canonical sphere.
Given a point $p_A$ on the source point cloud $S_A$, we denote its corresponding point on the target point cloud $S_B$ as $p_B$ and assume that they are decoded from two points $u_A$ and $u_B$ on the canonical sphere.
A one-layer linear operation can be written as:
\begin{equation}
     p_A = W\times u_A + b \quad\text{and}\quad p_B = W\times u_B + b 
     \label{eq:folding}
\end{equation} 
where $W$ and $b$ are learnable parameters instantiated as a fully-connected layer. 
Since Eq.~\eqref{eq:folding} shows an affine transform, it implies that $p_A - p_B \propto u_A - u_B$. If $u_A \neq u_B$ and $u_A$ is deformed to another point $p_B'$ on shape $S_B$ such that $p_A - p_B' < p_A - p_B$, which contradicts to the fact that $p_B$ is the closest point to $p_A$ among all points on $S_B$. Thus $u_A = u_B$, showing corresponding points are deformed from the same points on the canonical sphere.
%
%

%
%
%

\vspace{-2mm}
\paragraph{Inference via CPAE.}
\label{section:methodology:inference}
%
Given a source shape $S_{A}$ and a query point $p_A\in S_A$, we target at searching the correspondence of the query point in a target shape $S_{B}$.
We first compute the shape latent vectors for the two shapes as $z_A$ and $z_B$.
The query point is then mapped to a point $u_A$ in the canonical space by the canonical mapping encoder.
By feeding the concatenation of $u_A$ and $z_B$ to the inverse mapping decoder, we further map $u_A$ to a point $p_{A\rightarrow B}$ on the reconstructed target shape $S_{A\rightarrow B}$.
The correspondence of $p_A$ (denoted as $p_B\in S_B$) is thus the closet point on $S_{B}$ to $p_{A\rightarrow B}$.
%
%
%

\vspace{-2mm}
\paragraph{Confidence of Correspondence.} For each point and its correspondence pair (e.g. $(p_A, p_B)$), we can also compute a confidence score $C(p_A, p_B)$ to measure the confidence of this mapping:
\begin{equation}
C(p_A, p_B) = 1 - D(p_A, p_B)
\end{equation}
where $D$ refers to the normalized Euclidean distance between $p_A$ and $p_B$ in the 3D world coordinate. 
Similar to~\cite{liu2020learning}, if $C$ is lower than a pre-defined threshold $\tau$, we conclude that point $p_A$ does not have a correspondence on shape $S_B$.


\pdfoutput=1
\begin{figure*}[t]
\begin{center}
\includegraphics[width=1.0\linewidth]{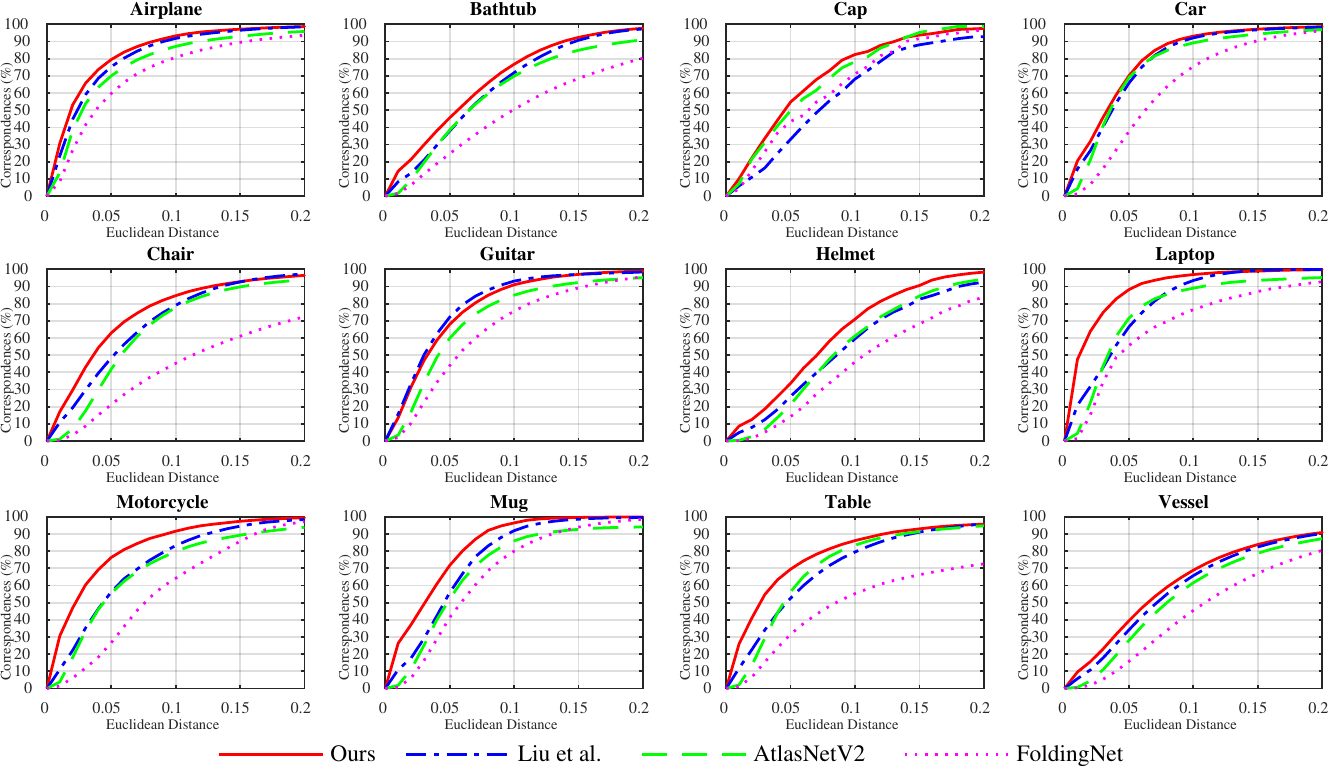}
\end{center}
\caption{Correspondence accuracy for 12 categories in the KeypointNet dataset.}
\label{fig.correspondence}
\end{figure*}

\vspace{-1mm}
\section{Experiments}
\vspace{-1mm}
\label{section:experiments}
In this section, we present evaluations of the proposed dense correspondence learning approach. 
%
%
To the best of our knowledge, there is no benchmark that provides ground-truth dense correspondences for general objects, which is exactly our motivation to learn dense correspondence with self-supervision. 
However, thanks to the learned dense correspondences across pairs of instances, we are able to carry out the task of 3D semantic keypoint transfer and part segmentation label transfer to evaluate the proposed method as in~\cite{liu2020learning}.
In the following, we first introduce our experimental setup as well as baselines.
%
We then report quantitative and qualitative comparisons with these baselines
for 3D semantic keypoint transfer and part segmentation label transfer. 
Finally, we present ablation studies to demonstrate the contribution of each component in the proposed model.
More results can be found in the appendix.
The source code will be made available to the public. 

\begin{figure*}[t]
\begin{center}
\includegraphics[width=1.0\linewidth]{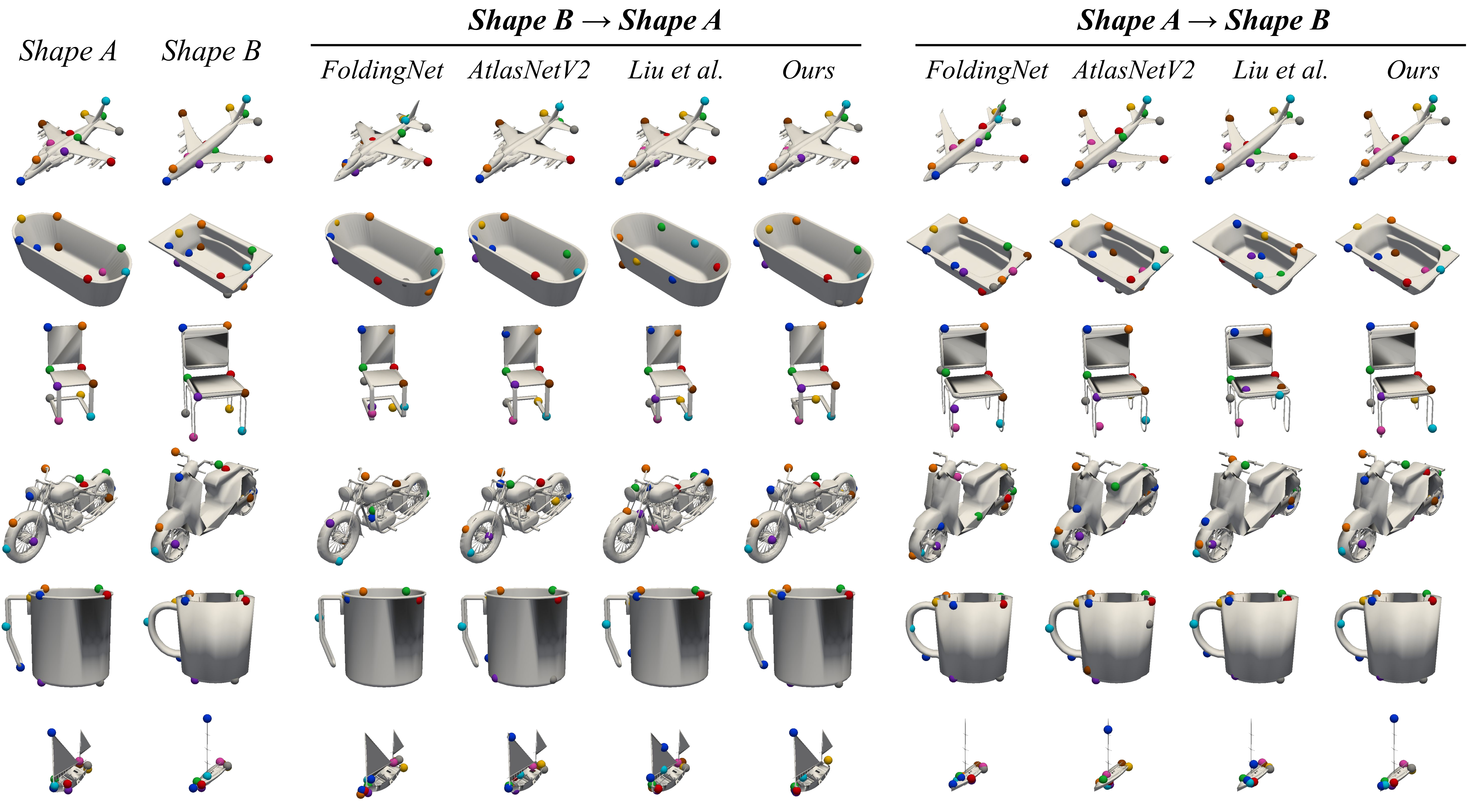}
\end{center}
\caption{Keypoint transfer results for five categories: airplane, bathtub, chair, motorcycle, mug, and vessel. Each row contains two shape each with ground-truth keypoints and its pairwise transfer result.}
\label{fig.qua.key}\vspace{-5mm}
\end{figure*}

\vspace{-1mm}
\subsection{Experimental Setup}
\label{section:experiments:setup}

\vspace{-2mm}
\paragraph{Dataset.}
We carry out the semantic keypoint transfer task on the KeypointNet dataset~\cite{you2020keypointnet} as the BHCP benchmark used in Liu et al.~\cite{liu2020learning} is not publicly available.
%
%
Compared to the BHCP benchmark, the KeypointNet dataset is more challenging because: (a) it contains diverse objects and comes with large-scale annotations, (b) it is template-free and annotated by a large group of people, thus is less biased compared to the keypoints in the BHCP benchmark, which are from predefined templates.
For the part segmentation label transfer task, we use the ShapeNet part dataset~\cite{yi2016scalable} as in~\cite{liu2020learning}.
For both datasets, we use the split provided in the original paper, and generate all pairs of shapes in the testing set as our testing pairs. 
To avoid interference from non-existing correspondences, we leave out instance pairs that do not share the same keypoint or part label.
In all experiments, including our method and the baselines, we use a validation set for model selection.

\vspace{-2mm}
\paragraph{Baselines.}
We evaluate the proposed method against state-of-the-art learning-based 3D dense correspondence prediction approaches, including AtlasNetV2~\cite{deprelle2019learning}, FoldingNet~\cite{yang2018foldingnet}, and Liu et al.~\cite{liu2020learning}.
%
%
Specifically, FoldingNet deforms from a fixed UV grid, while AtlasNetV2 explicitly allows the shape to be deformed from learnable elementary 3D structures. 
Neither of them estimates the confidence of correspondences. 
%
Liu et al.~\cite{liu2020learning} propose a method that utilizes part features learned by the BAE-NET~\cite{chen2019bae} to learn dense correspondence, including a mechanism that estimates the correspondence confidence.

\vspace{-2mm}
\paragraph{Implementation Details.}
%
For both the point canonical mapping encoder and point inverse mapping decoder (see Figure~\ref{fig.overview}), we follow~\cite{deprelle2019learning} and use a three-layer MLP with ReLU activations, BatchNorm layers except for the last layer, where we use a hyperbolic tangent activation to obtain the final output. 
%
%
The training phase of our approach consists of two stages: (1) A pre-training stage trained with $L_{ACD}$ (Eq.~\ref{eq:unfolding}) and $L_{rec}$ (Eq.~\ref{eq:Rec}) using $\alpha=1$ for $L_{ACD}$ (2) A fine-tuning stage trained with $L_{ACD}$, $L_{rec}$, and $L_{cross}$ (Eq.~\ref{eq:cross}) where we set $\alpha=0$ for $L_{ACD}$.
For all experiments, we set $k=2048$, $\tau=0.9$ (see Section~\ref{section:methodology:inference}), and the parameters of the network are optimized using the Adam~\cite{kingma2013auto} optimizer, with a constant learning rate of $1e^{-4}$.

\vspace{-1mm}
\subsection{3D Semantic Keypoints Transfer}
\vspace{-1mm}
\label{section:experiments:keypoint}
For fair comparisons, we follow~\cite{liu2020learning,chen2020unsupervised} and compute the distances from transferred keypoints to ground truth keypoints and report the percentage of testing pairs where the distances are below a given threshold in Figure~\ref{fig.correspondence}.
We demonstrate that for 11 out of 12 categories (e.g. airplane, chair, mug, etc.), keypoints transferred via our learned correspondence are more accurate than other methods~\cite{liu2020learning,deprelle2019learning,yang2018foldingnet}. 
At the distance threshold of 0.05, our method performs 11.2\% more accurately than Liu et al.~\cite{liu2020learning} on average of all categories.
Figure~\ref{fig.correspondence} demonstrates the qualitative results of the keypoint transfer task. Even for categories with large intra-class variation, e.g. bathtub, motorcycle, or vessel, our method is able to transfer keypoints accurately thanks to the learned dense correspondence. 



\begin{table*}[t]
\centering
{
\tiny
\begin{tabular}{lllllllllllllllll}
\hline
 & pla. & bag & cap & cha. & ear. & gui. & kni. & lam. & lap. & bik. & mug & pis. & roc. & ska. & tab. & avg. \\ \hline
 \hline
Liu et al. & 60.1 & 56.2 & 59.7 & 72.2 & 45.3 & \bf{81.5} & 66.4 & 42.6 & 88.5 & 40.5 & \bf{87.5} & \bf{66.4} & 37.2 & 50.7 & 70.4 & 61.7 \\ \hline
Ours & \bf{61.3} & \bf{59.3} & \bf{61.6} & \bf{72.6} & \bf{55.5} & 78.9 & \bf{71.3} & \bf{53.2} & \bf{89.9} & \bf{55.4} & 86.5 & 66.2 & \bf{40.2} & \bf{61.8} & \bf{72.5} & \bf{65.8}   \\ \hline
\end{tabular}
}
\caption{Part label transfer results for 15 categories in the ShapeNet part dataset. Number measured with average IOU(\%).}
\label{tab.part}\vspace{-5mm}
\end{table*}

\vspace{-1mm}
\subsection{Part Segmentation Label Transfer}
\begin{figure*}[t]
\begin{center}
\includegraphics[width=1.0\linewidth]{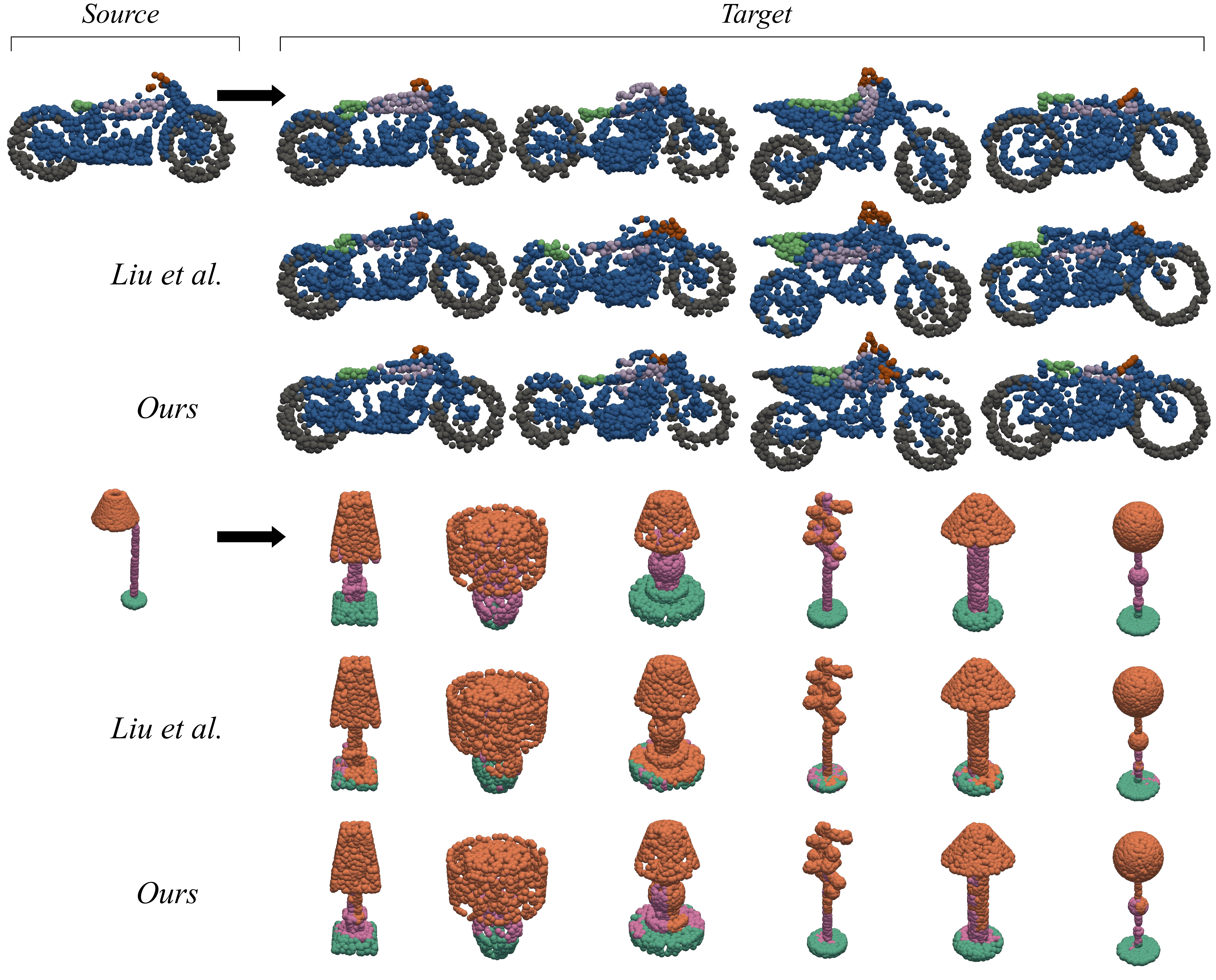}
\end{center}
\caption{Qualitative results of part label transfer. The 1st row of the target indicates the ground truth instance part labels, while the rest shows the label transferring results via learned correspondences.}
\label{fig.qua.part}\vspace{-3mm}
\end{figure*}

\vspace{-1mm}
\label{section:experiments:part}
We further validate our approach on the part label transfer task and present quantitative results in Table~\ref{tab.part} and qualitative results in Figure~\ref{fig.qua.part}.
Note that the settings are different from that of Liu et al.~\cite{liu2020learning}, which directly utilizes the branched co-segmentation results for evaluation (thus different quantitative results between the Table~\ref{tab.part} and that in their paper).
In Table~\ref{tab.part}, we show the intersection-over-union (IoU) between transferred and ground truth part labels. 
Our method performs better than Liu et al.~\cite{liu2020learning} in 12 out of 15 categories and has a higher average IoU.
For categories with large intra-class variations, such as lamps, motorbikes, our approach significantly outperforms Liu et al.~\cite{liu2020learning}. 
%
The performance difference is most likely due to the branched co-segmentation~\cite{chen2019bae} adopted by Liu et al.~\cite{liu2020learning}. 
Branched co-segmentation relies heavily on the size of training data available. Such an approach is vulnerable to large shape variations~\cite{chen2019bae}, and is unable to segment flat surfaces. Moreover, its performance is sensitive to the pre-defined number of branches.  Instead, our method naturally links different instances through a canonical primitive and is thus more robust to large shape variations.
We show the qualitative results in Figure~\ref{fig.qua.part}. Thanks to the dense correspondence learned by the proposed method, we are able to transfer small parts more accurately such as the seat of a motorcycle (the green points) and the pipe of a lamp (the purple points) comparing to Liu et al.~\cite{liu2020learning}.
%

\begin{figure}[t]
\begin{center}
\includegraphics[width=1.0\linewidth]{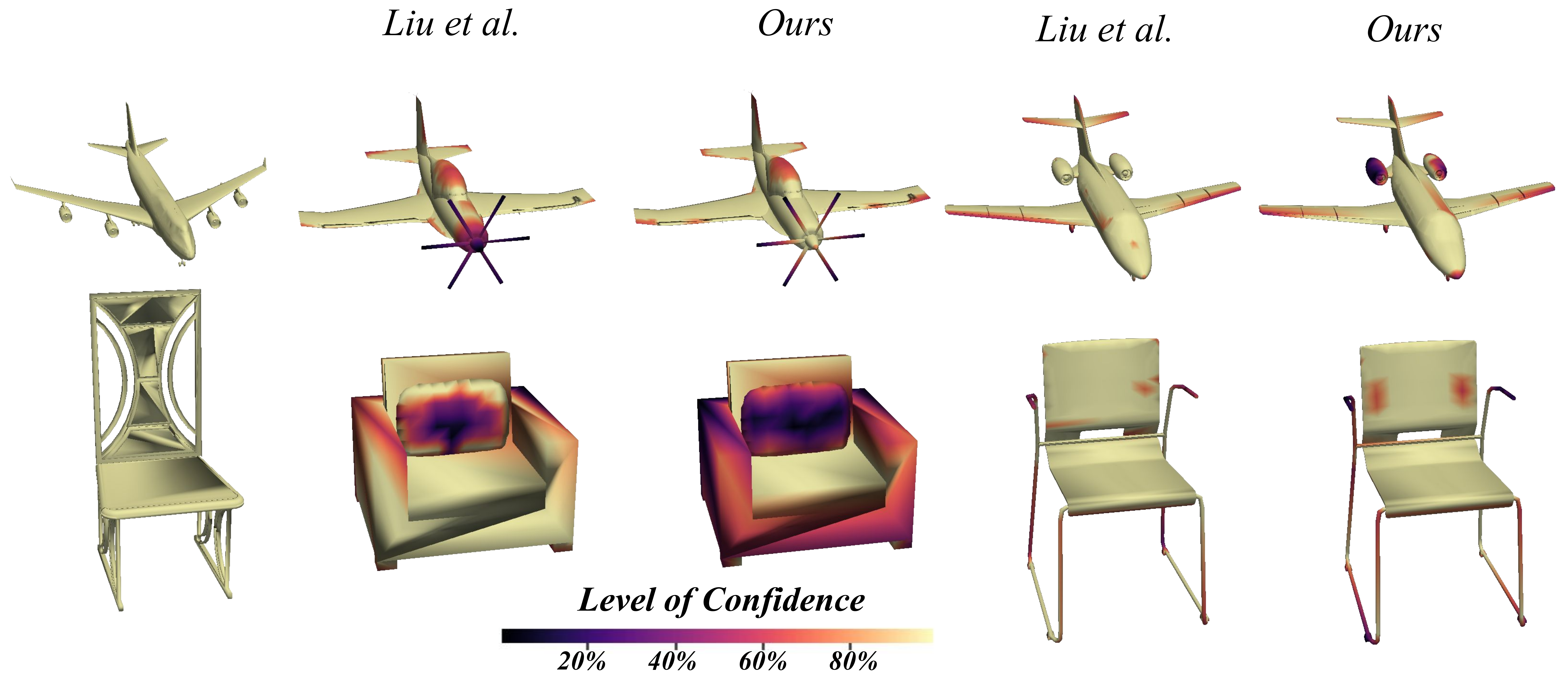}
\end{center}
\caption{Heatmaps representing the correspondence confidence generated by our network. Source shape is in the leftmost of each row and dark color in each heatmap refers to low confidence in existing correspondence.}
\label{fig.heatmap}
\end{figure}

\vspace{-1mm}
\subsection{Correspondence Confidence}
\vspace{-1mm}
\label{section:experiments:existence}
Given a source shape $S_{A}$ and a target shape $S_{B}$, for every point $q \in S_{B}$, our model computes a confidence score indicating whether its corresponding point exists in $S_{A}$, as discussed in Section~\ref{section:methodology:inference}.
As annotations can be biased by the annotator's definitions on keypoints, there is no absolute ground-truth for non-existence label between a shape pair.
Thus, we visualize the confidence score as heatmaps for multiple target shapes in the airplane category in Figure~\ref{fig.heatmap} and compare with Liu et al.~\cite{liu2020learning}.
%
%
%
%
%
%
As shown in Figure~\ref{fig.heatmap}, the proposed method is able to produce a more fine-grained confidence score compared to Liu et al.~\cite{liu2020learning}.
%
This is because our approach explicitly evaluates the confidence of correspondences at a more fine-grained level -- the distance between points, instead of a distance at the part-level, as proposed in~\cite{liu2020learning}.


\begin{figure}[ht]
\begin{subfigure}{.5\textwidth}
  \centering
    \includegraphics[width=1\textwidth]{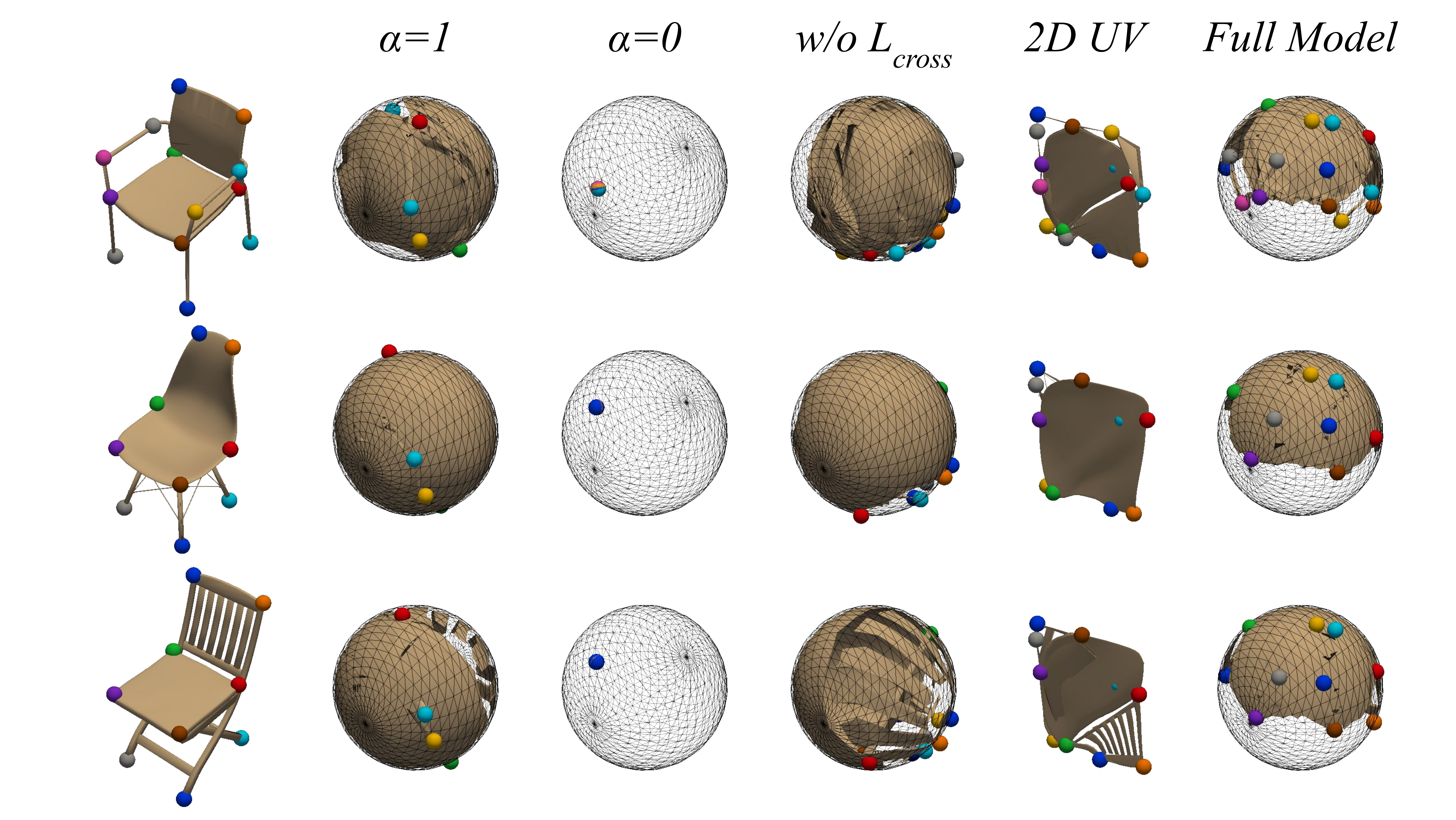}
   \caption{}
  \label{fig:abl-qual}
\end{subfigure}
\begin{subfigure}{.5\textwidth}
  \centering
    \includegraphics[width=1\textwidth]{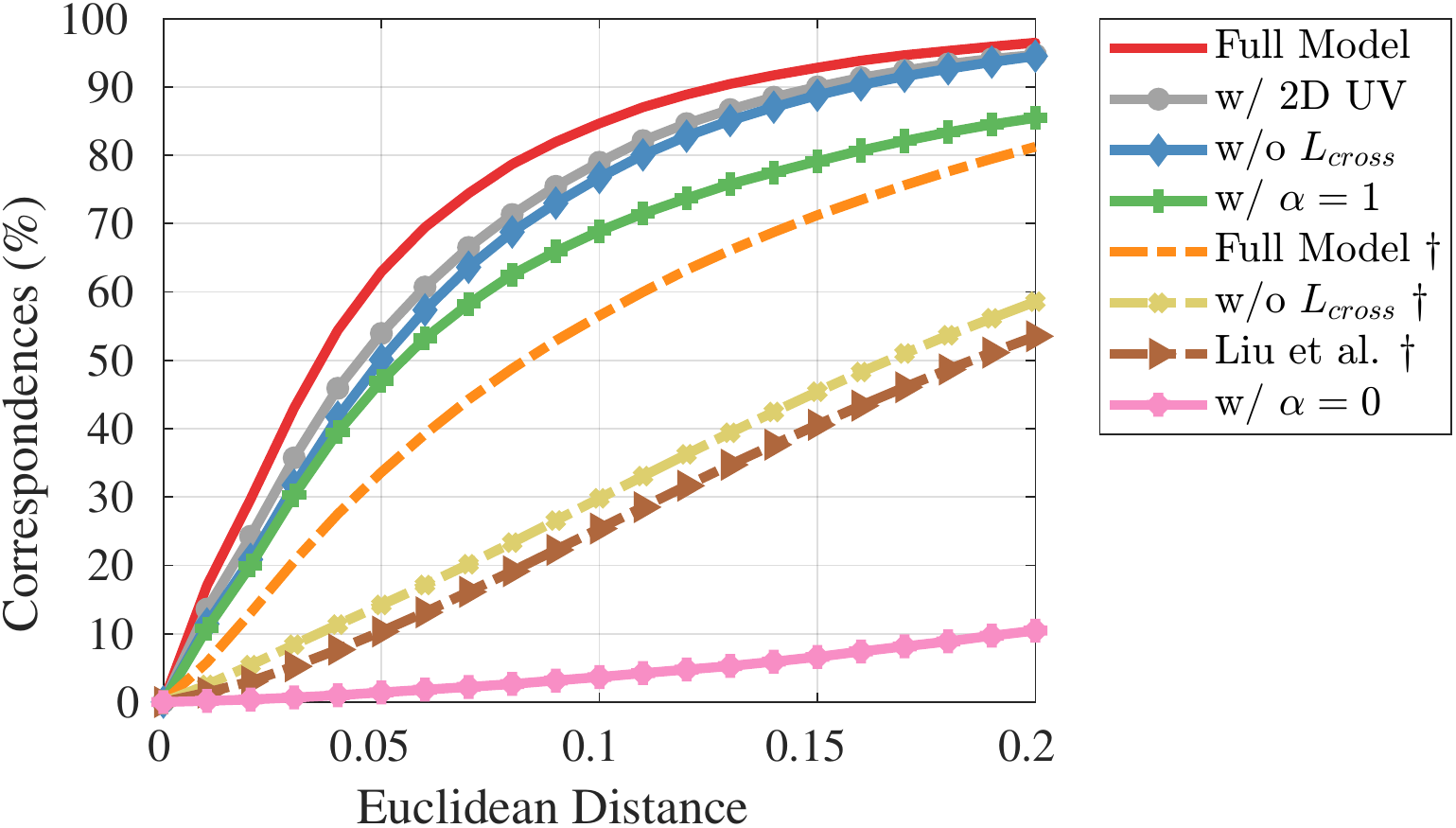}
   \caption{}
  \label{fig:abl-quan}
\end{subfigure}
\caption{Qualitative (a) and quantitative (b) results of ablation studies on (i) the cross-reconstruction loss term (ii) the adaptive Chamfer loss term (iii) different types of canonical UV primitive using the chair category in the KeypointNet dataset. Note that the $\dagger$ sign indicates methods trained with an un-aligned setting.}
\label{fig:abl}\vspace{-3mm}
\end{figure}

\vspace{-1mm}
\subsection{Ablation Studies}
\label{section:experiments:ablations}

\vspace{-2mm}
\paragraph{Effectiveness of cross-reconstruction.}
The cross-reconstruction architecture is designed to ensure that corresponding points overlapping as much as possible on the canonical space. 
In our experiments, we find that with the datasets containing shapes with aligned 6D poses~\cite{you2020keypointnet, chang2015shapenet}, a single branch encoder-decoder framework (i.e., w/o $L_{cross}$) already produces a reasonable prediction of correspondences. However, as shown in Figure~\ref{fig:abl} (red vs blue), our cross-reconstruction framework significantly improves the performance with a large margin.

In addition, we validate the effectiveness of the framework on un-aligned shapes, by rotating the input point cloud with radian noise $\mathcal{N}(0,\,0.5^{2})$. We note that all co-segmentation approaches assume that the pose of input shapes are consistently aligned~\cite{chen2019bae, liu2020learning}. Consistent with the assumption, such rotation severely degrades the performance of Liu et al.~\cite{liu2020learning} (brown line). In contrast, with the help of cross-reconstruction loss (orange line), our model is rotation-invariant to a certain degree.


\vspace{-2mm}
\paragraph{Effectiveness of adaptive Chamfer loss.}
To show that the adaptive Chamfer loss is necessary, we train our model with constant $\alpha=0$ and $\alpha=1$, respectively. 
The results are shown in Figure~\ref{fig:abl}. 
When $\alpha$ is set to zero consistently, the primitive would eventually condense to a single point in the canonical space, therefore, significantly hurts the performance (pink line). 
On the other hand, if $\alpha=1$, it is equivalent to enforce a single canonical primitive for all instances by encouraging the primitive to cover the entire canonical space.
This ignores the intra-class variation and deteriorates the performance as shown in Figure~\ref{fig:abl} (green line).

\vspace{-2mm}
\paragraph{Using a 2D UV grid vs. 3D UV sphere as a primitive.}
To analyze the effect of different canonical UV primitives, we further train our model using a 2D UV grid instead of a 3D UV sphere. 
Figure~\ref{fig:abl} shows that parameterizing the canonical space with a 3D sphere quantitatively outperforms a 2D grid (grey line).
The main reason is that the sphere is continuous at any point while a 2D grid is discontinuous at boundaries.

\pdfoutput=1
\vspace{-1mm}
\section{Conclusions}
\vspace{-1mm}
In this work,  we propose a self-supervised model, CPAE, that learns dense correspondence between 3D shapes in the same category.
We introduce a canonical UV sphere, where dense correspondence for all the shapes can be explicitly obtained from.
The key is to learn a 3D world coordinate to canonical space mapping, so that points from different instances are regarded as a pair of correspondences if they overlap on the sphere.
We fulfill it through an autoencoder equipped with an adaptive Chamfer loss in the bottleneck, and a cross-reconstruction structure in the decoder. 
Experimental results validate the proposed method performs favorably against state-of-the-art schemes in  various tasks and ablations. 
We show that our model is much more rotation-invariant than the existing approaches.

\newpage

\newpage
\pdfoutput=1
\section*{\Large Appendix}

In the Appendix, we provide:

\begin{enumerate}
    \item[$\diamond$] Application on texture transfer in Section~\ref{sup:texture}.
    \item[$\diamond$] The computational time and model size of our approach in Section~\ref{sup:computational}.
    \item[$\diamond$] Qualitative results of instance-aware primitives in Section~\ref{sup:primitives}.
    \item[$\diamond$] Qualitative results of part label transfer in Section~\ref{sup:part}.
    \item[$\diamond$] Correspondence Confidence Heatmap Visualization in Section~\ref{sup:confidence}.
    \item[$\diamond$] Limitation and failure cases in Section~\ref{sup:limitation}.
\end{enumerate}

\section{Application on Texture Transfer}
\label{sup:texture}
Given a source 3D shape, we transfer texture from the source shape to multiple target shapes using computed correspondences. Our method is able to detect points that do not have correspondents in the source 3D shape (e.g., airplane without tail wings or stabilizers).

\begin{figure}[h]
\begin{subfigure}{.26\textwidth}
  \centering
  \includegraphics[width=.9\linewidth]{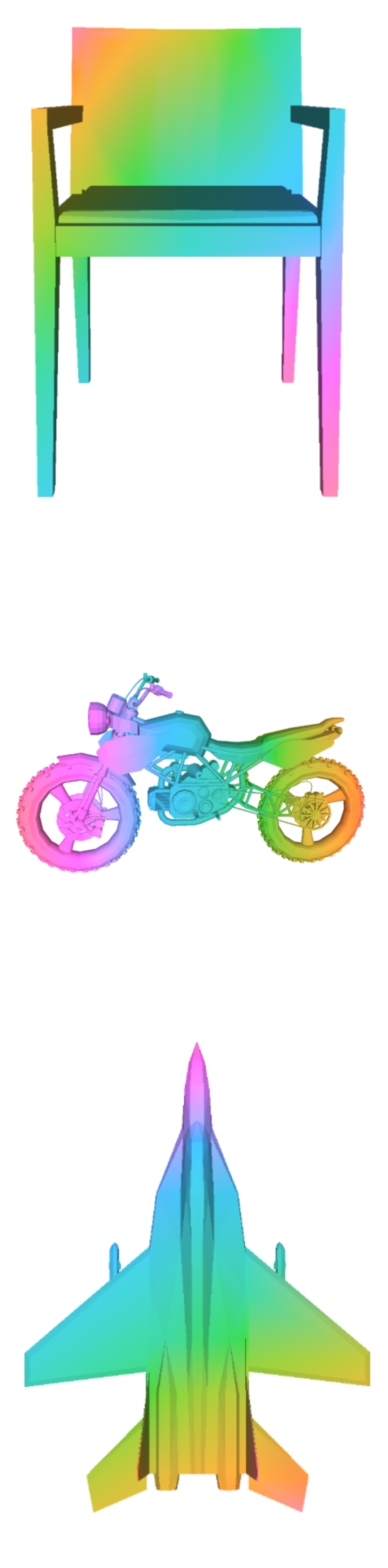}
  \caption{Source}
  \label{fig:sub-first}
\end{subfigure}
\begin{subfigure}{.8\textwidth}
  \centering
  \includegraphics[width=.8\linewidth]{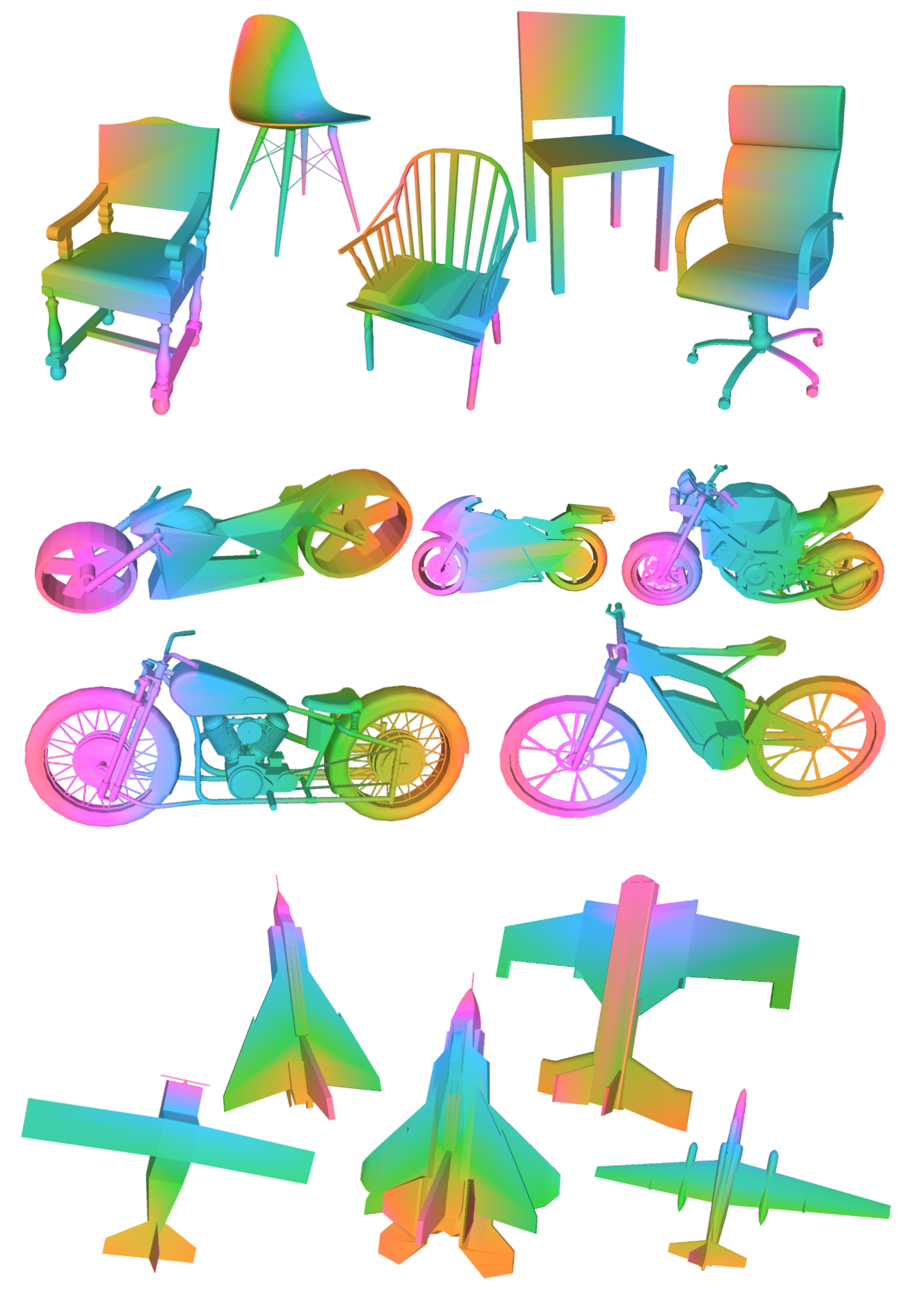}
  \caption{Targets}
  \label{fig:sub-second}
\end{subfigure}
\caption{Applications on texture transfer.}
\label{fig:fig}
\end{figure}

\newpage

\section{Computational Time and Model Size}
\label{sup:computational}

\vspace{-2mm}
\subsection{Training Details}
We use the chair category as an example to illustrate the training pipeline.
With 500 training samples, the training takes about 24 hours to converge, (4 hours for stage one ($\alpha=1$) and 20 hours for stage two ($\alpha=0$)), on a single Tesla V100 GPU. 
Please see Algorithm~\ref{pseudo} and Section 4.1 in the main paper for more details on the two stage training paradigm.
\textit{The source code will be released to public upon publication.}

\vspace{-2mm}
\begin{algorithm}[H]
    
    \caption{\textbf{:} The training phase of our approach consists of two stages: (1) A pre-training stage trained with $L_{ACD}$ and $L_{rec}$ (2) A fine-tuning stage trained with $L_{ACD}$, $L_{rec}$, and $L_{cross}$}
    \label{algo:overview}
    \vspace{0.2em}
    \begin{algorithmic}
        \STATE \hspace{-0.8em} \underline{\sc{\textbf{(A) Stage-1: Pre-training}}} \textit{\small \hfill $\vartriangleright$ \underline{4 hours on Chair category}}
    \end{algorithmic}
    \vspace{0.1em}
    \begin{algorithmic}[1]
        \STATE \hspace{0.5em} Randomly sub-sample $k$ points from the input point cloud $S_A$;
        \STATE \hspace{0.5em} Initialize weight of the global feature encoder $E(\cdot)$, canonical mapping encoder $\Phi(\cdot)$ and inverse mapping decoder $\Psi(\cdot)$;
        
        \STATE \hspace{0.5em} \textbf{for} {epoch} \textbf{in} \textbf{range} [0,T) \textbf{do}
        \STATE \hspace{1.5em} \textbf{foreach} {iteration} \textbf{do}
        \STATE \hspace{2.5em} $\boldsymbol{z_A} \leftarrow \, E(S_A)$;
        \STATE \hspace{2.5em} $\boldsymbol{U_A} \leftarrow \, \Phi([p,z_A])$, where $p \in S_A$;
        \STATE \hspace{2.5em} $\boldsymbol{\hat{S}_{A \rightarrow A}} \leftarrow \, \Psi([q,z_A])$, where $q \in U_A$;
        \STATE \hspace{2.5em} Obtain loss $L_{ACD}$ ($\alpha=1$) and $L_{rec}$;
        \STATE \hspace{2.5em} Update weight;
        
    \end{algorithmic}
    \vspace{-0.7em} \hrulefill \\ [-1.2em]
    \begin{algorithmic}
        \STATE \hspace{-0.8em} \underline{\sc{\textbf{(B) Stage-2: Fine-Tuning}}} \textit{\small \hfill $\vartriangleright$ \underline{20 hours on Chair category}}
    \end{algorithmic}
    \vspace{0.1em}
    \begin{algorithmic}[1]
        \STATE \hspace{0.5em} Generate randomly paired samples $S_A$ and $S_B$;
        \STATE \hspace{0.5em} \textbf{while} not converged \textbf{do}
        \STATE \hspace{1.5em} \textbf{foreach} {iteration} \textbf{do}
        \STATE \hspace{2.5em} $\boldsymbol{z_A} \leftarrow \, E(S_A)$;
        \STATE \hspace{2.5em} $\boldsymbol{U_A} \leftarrow \, \Phi([p,z_A])$, where $p \in S_A$;
        \STATE \hspace{2.5em} $\boldsymbol{\hat{S}_{A \rightarrow A}} \leftarrow \, \Psi([q,z_A])$, where $q \in U_A$;
        \STATE \hspace{2.5em} Obtain loss $L_{ACD}$ ($\alpha=0$) and $L_{rec}$;
        \STATE \hspace{2.5em} $\boldsymbol{z_B} \leftarrow \, E(S_B)$;
        \STATE \hspace{2.5em} $\boldsymbol{U_B} \leftarrow \, \Phi([p,z_B])$, where $p \in S_B$;
        \STATE \hspace{2.5em} $\boldsymbol{\hat{S}_{A \rightarrow B}} \leftarrow \, \Psi([q,z_B])$, where $q \in U_A$;
        \STATE \hspace{2.5em} $\boldsymbol{\hat{S}_{B \rightarrow A}} \leftarrow \, \Psi([q,z_A])$, where $q \in U_B$;
        \STATE \hspace{2.5em} Obtain loss $L_{cross}$;
        \STATE \hspace{2.5em} Update weight;
    \end{algorithmic}
    \label{pseudo}
\end{algorithm}

\begin{wrapfigure}{r}{0.5\textwidth}
     \vspace{-15pt}
     \centering
     \includegraphics[width=0.48\textwidth]{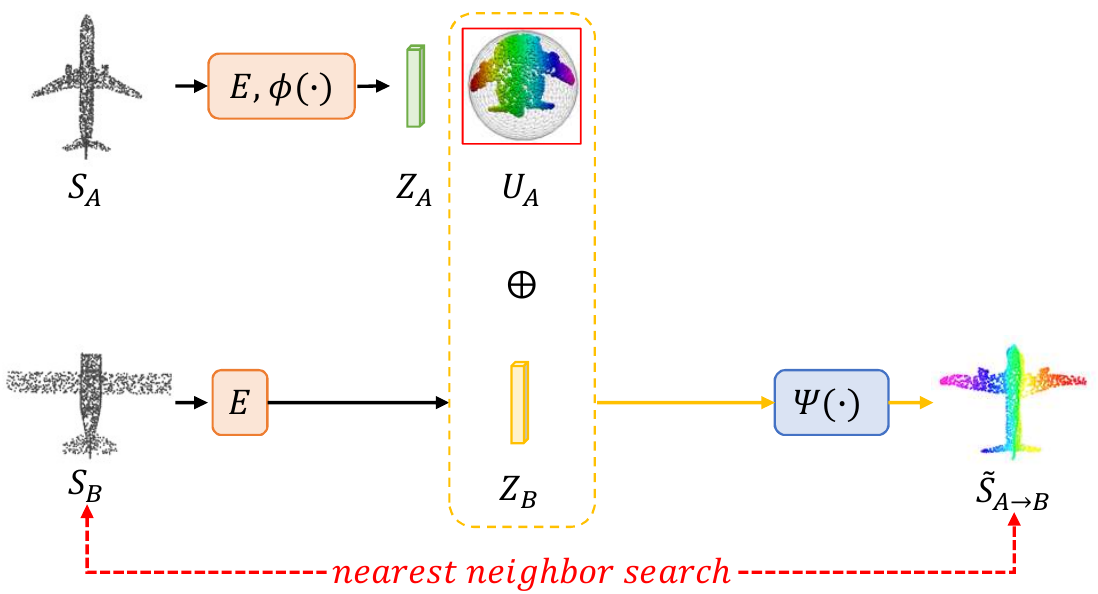}
     \caption{The inference flow of our approach.}
     \label{fig.inference}
     \vspace{-15pt}
 \end{wrapfigure}

\subsection{Inference Details}
Please see Figure~\ref{fig.inference} and Section 3.2 in the main paper for more details on the inference pipeline.
Our proposed CPAE model contains 2.07M parameters which is 2.5$\times$ less than the 5.22M parameters in~\cite{liu2020learning}.
At inference time, the computational time for label transfer between a pair of shapes (each with 2048 points) is 0.03 second including runtimes of the nearest neighbor search for both shapes. 

\clearpage

\section{Qualitative Results of Instance-aware Primitives}
\label{sup:primitives}
\begin{figure}[h]
\begin{center}
\includegraphics[width=1\linewidth]{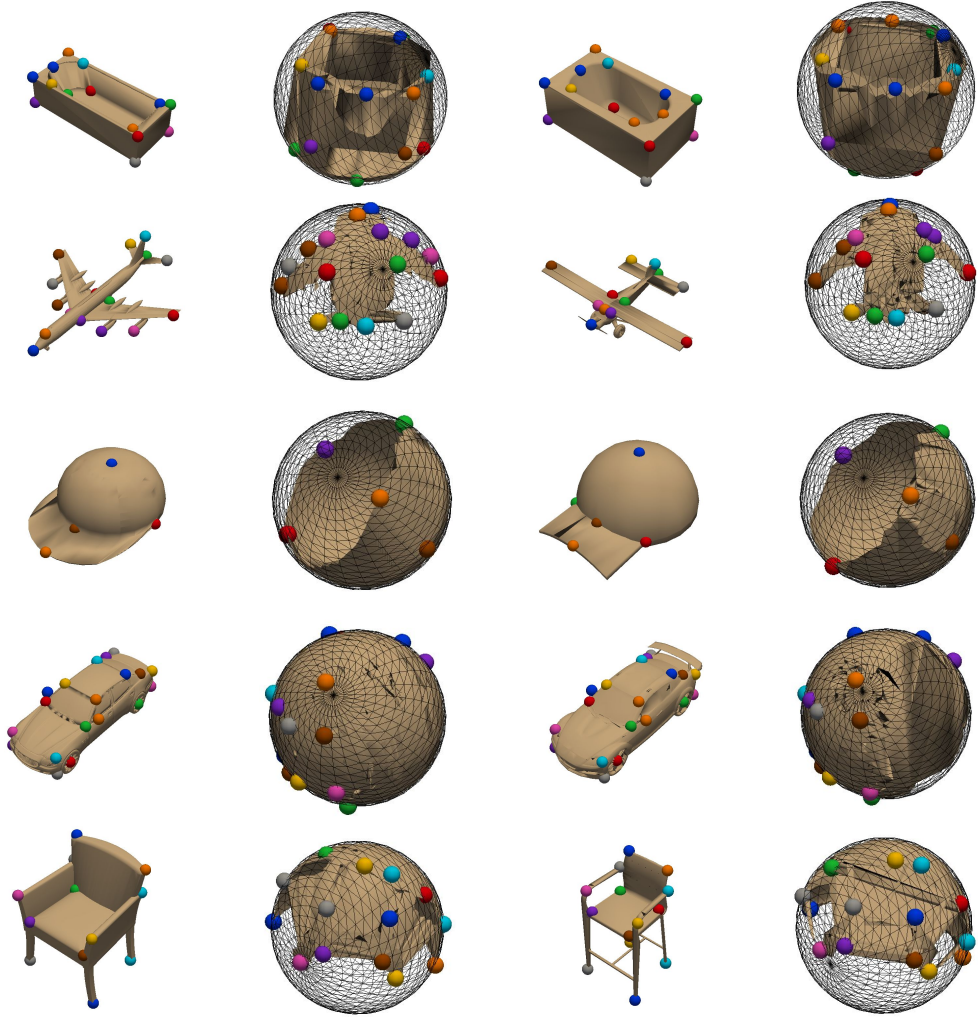}
\end{center}
\caption{Iinstance-aware primitives on five different categories from the KeypointNet dataset~\cite{you2020keypointnet}}
\label{fig.supp.prim}
\end{figure}

In Figure~\ref{fig.supp.prim}, we demonstrate more examples of the instance-aware primitives as discussed in Section 3.1 in the main paper. 
Two observations can be made from this figure: a) The instance-aware primitives produced by our canonical mapping are closely adhered to a canonical sphere. b) points of the same semantic parts are mapped to nearby locations on the primitives, as shown by the colored keypoints in Figure~\ref{fig.supp.prim}. These observations demonstrate that our model is able to learn correspondence across different shapes in the same category.

\newpage

\section{Qualitative Results of Part Label Transfer}
\label{sup:part}
\begin{figure}[h]

\begin{center}
\includegraphics[width=0.95\linewidth]{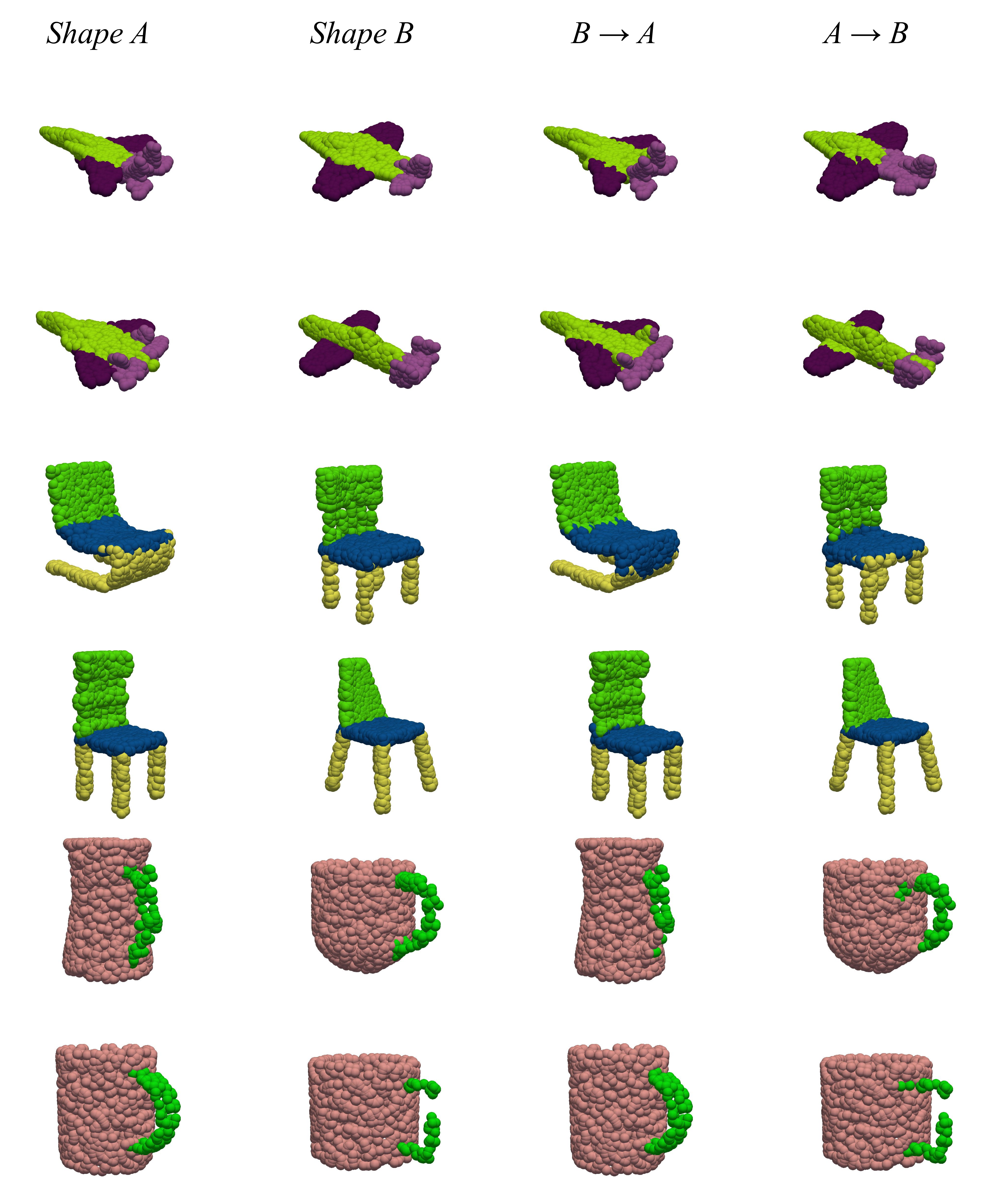}
\end{center}
\caption{Part label transfer results. B$\rightarrow$A refers to transferring shape B's label to shape A.}
\label{fig.supp.part}
\end{figure}

In Figure~\ref{fig.supp.part}, we show more qualitative results of the part label transfer task. Thanks to the dense correspondences learned by our model, we can transfer part labels for small parts (e.g. the handle of mugs or the tail wings of airplanes) and handle large intra-class variations (e.g. different legs of chairs in row three in Figure~\ref{fig.supp.part}).

\newpage

\section{Correspondence Confidence Heatmap Visualization}
\label{sup:confidence}
\begin{figure}[h]
\begin{center}
\includegraphics[width=1\linewidth]{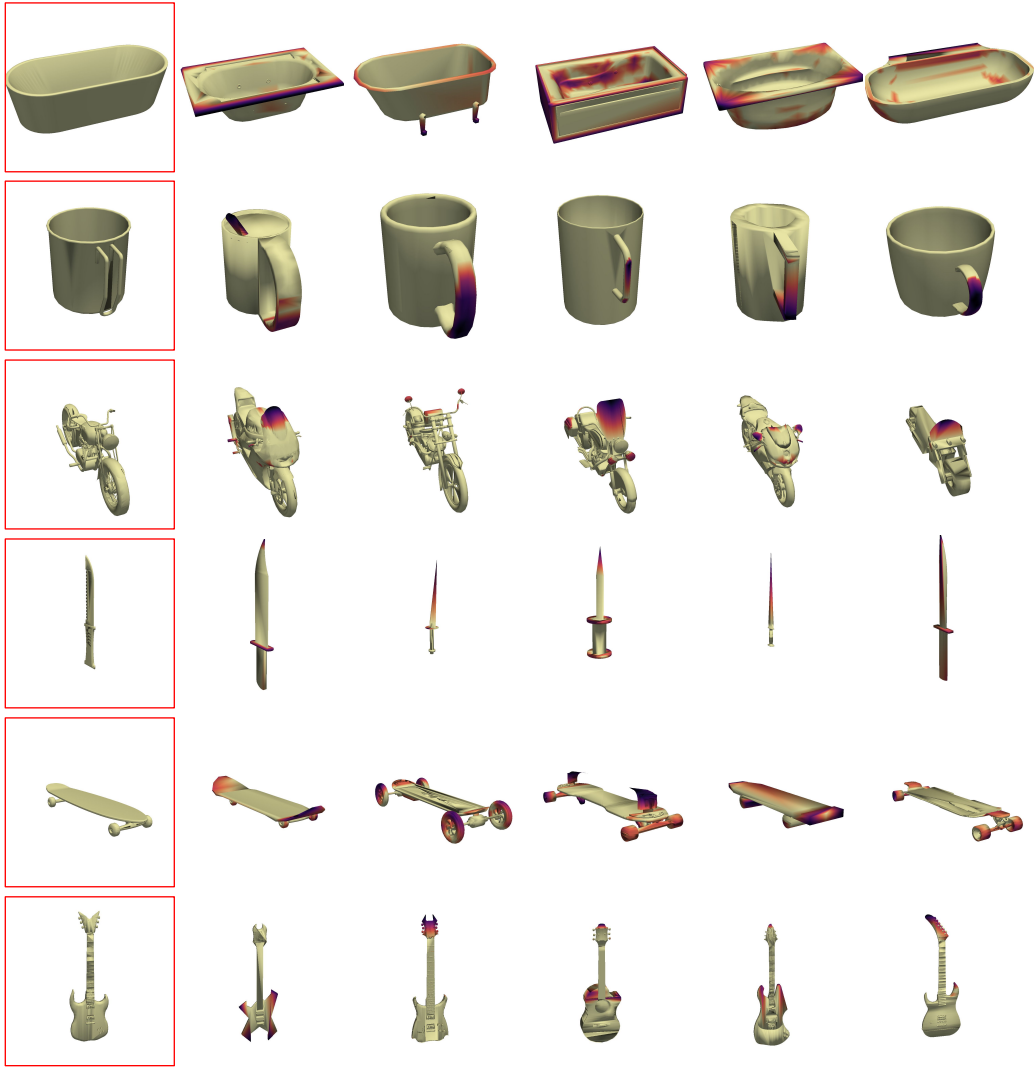}
\end{center}
\caption{Correspondence confidence heatmaps. Red boxes indicate source shapes. The darker the heatmap, the lower the confidence. 
}
\label{fig:conf}
\end{figure}

We use the confidence score (mentioned in main paper Section 3.4) to draw heatmaps for multiple target shapes in the same categories. 
As shown in Figure~\ref{fig:conf}, the predicted confidence heat maps successfully indicate the intra-class variations and capture uncertainty in correspondence prediction. 
For instance, low confidences can be found in mugs with different handles, knives with different blades, guitars with different bodies, etc.

\newpage

\section{Limitation}
\label{sup:limitation}
There are two main limitations for our approach: 
(a) we encode the shape information of a point cloud in a global vector -- i.e., fine details like corners and edges may be blurred after reconstruction.
%
(b) We found the correspondences predicted near holes maybe wrong, possibly due to the sparsity of points in the point cloud and the nature of Chamfer and Earth Mover's distance matrices. We leave these limitations for future works.

\begin{figure}[h]
\begin{center}
\includegraphics[width=1\linewidth]{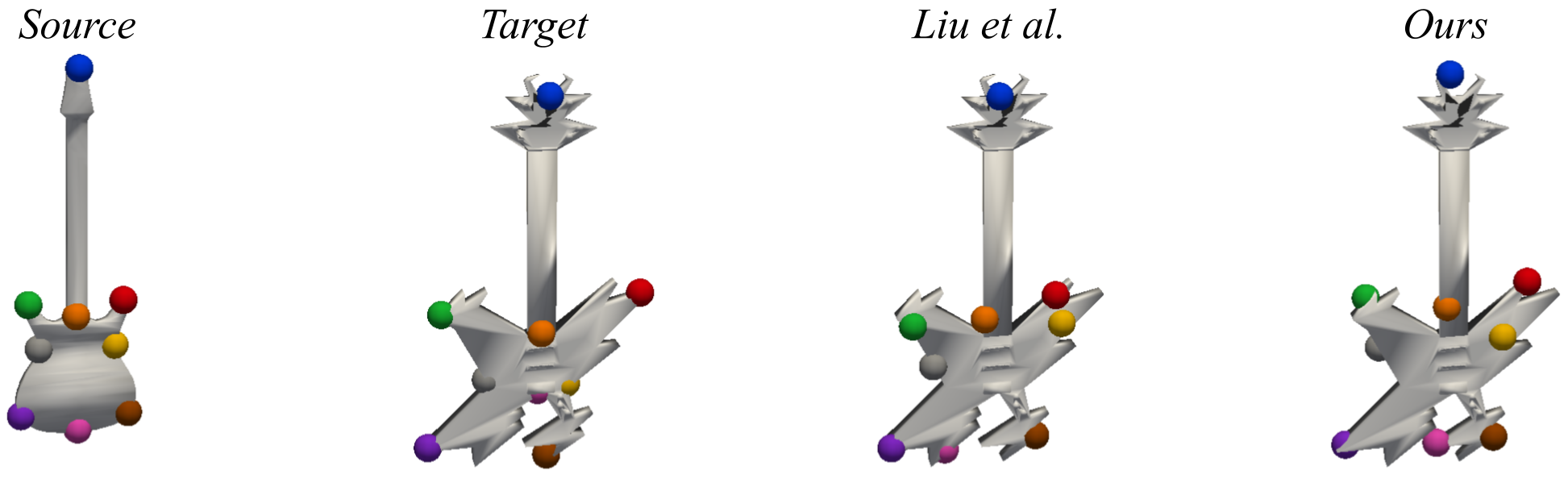}
\end{center}
\caption{Failure cases.}
\end{figure}

\clearpage
\bibliographystyle{unsrt}
\bibliography{main}

\end{document}